\documentclass[11pt,a4paper]{article}
\usepackage[hyperref]{naaclhlt2019}
\usepackage{times}
\usepackage{latexsym}
\usepackage{amsmath}
\usepackage{amssymb}
\usepackage{tabularx}
\usepackage{url}
\usepackage{booktabs}
\usepackage{graphicx}
\usepackage{multirow}
\usepackage{todonotes}
\renewcommand{\vec}[1]{\mathbf{#1}}

\aclfinalcopy 

\title{Tracking Discrete and Continuous Entity State for\\ Process Understanding}

\author{Aditya Gupta \and Greg Durrett \\
  Department of Computer Science \\
  The University of Texas at Austin \\
  {\tt \{agupta,gdurrett\}@cs.utexas.edu} }

\date{}

\begin{document}
\maketitle
\begin{abstract}
Procedural text, which describes entities and their interactions as they undergo some process, depicts entities in a uniquely nuanced way. First, each entity may have some observable discrete attributes, such as its state or location; modeling these involves imposing global structure and enforcing consistency. Second, an entity may have properties which are not made explicit but can be effectively induced and tracked by neural networks. In this paper, we propose a structured neural architecture that reflects this dual nature of entity evolution. The model tracks each entity recurrently, updating its hidden continuous representation at each step to contain relevant state information. The global discrete state structure is explicitly modelled with a neural CRF over the changing hidden representation of the entity. This CRF can explicitly capture constraints on entity states over time, enforcing that, for example, an entity cannot move to a location after it is destroyed.  We evaluate the performance of our proposed model on QA tasks over process paragraphs in the \textsc{ProPara} dataset \cite{dalvi2018tracking} and find that our model achieves state-of-the-art results.
\end{abstract}

\begin{figure*}[t!]
  \includegraphics[width=.87\textwidth]{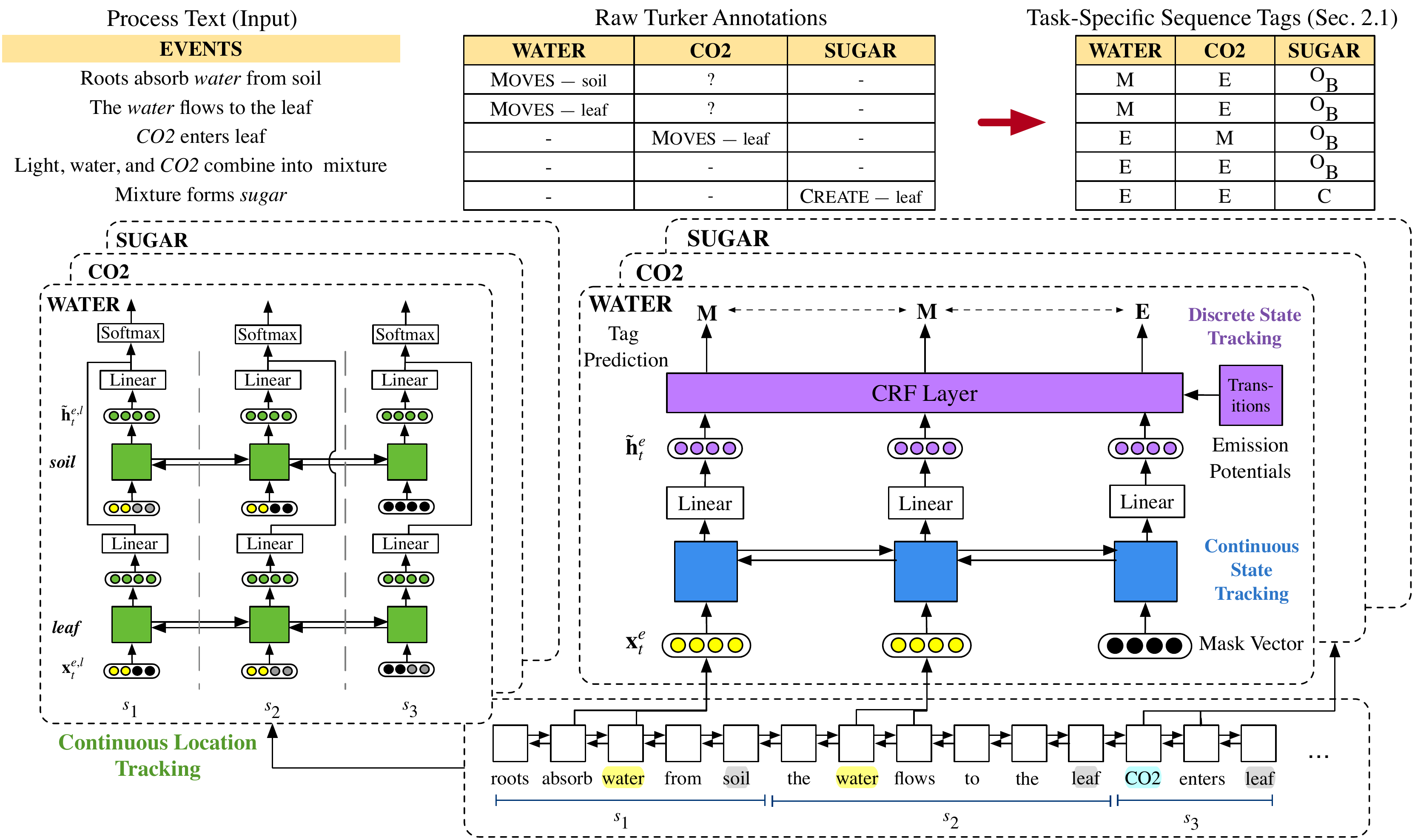}
  \centering
  \caption{Task and our proposed model. Top: raw text descriptions are annotated with entity-state change information; we modify this in a rule-based way for our model. Bottom: our model. Entity mention and verb information is aggregated in per-entity LSTMs (right). A CRF layer then predicts entity state. A separate sentence-level LSTM (left) tracks each entity-location pair using the combined entity and location mention information.  }
  \label{fig:grid}
\end{figure*}

\section{Introduction}

Many reading comprehension question answering tasks  \cite{richardson-burges-renshaw:2013:EMNLP,rajpurkarEtAlSquad,joshi-EtAl:2017:Long} require looking at primarily one point in the passage to answer each question, or sometimes two or three \cite{yangetal-Hotpot,welbletal-Wikihop}. As a result, modeling surface-level correspondences can work well \cite{seo2016query} and holistic passage comprehension is not necessary. However, certain QA settings require deeper analysis by focusing specifically on entities, asking questions about their states over time \cite{weston2015towards, long2016simpler}, combination in recipes \cite{bosselut2018simulating}, and participation in scientific processes \cite{dalvi2018tracking}. These settings then suggest more highly structured models as a way of dealing with the more highly structured tasks. One crucial aspect of such texts is the way an entity's state evolves with both discrete (observable state and location changes) and continuous (changes in unobserved hidden attributes) phenomena going on. Additionally, the discrete changes unfold in a way that maintains the state consistency: an entity can not be \textit{destroyed} before it even starts to \textit{exist}.

In this work, we present a model which both recurrently tracks the entity in a continuous space while imposing discrete constraints using a conditional random field (CRF). We focus on the scientific process understanding setting introduced in \citet{dalvi2018tracking}. For each entity, we instantiate a sentence-level LSTM to distill continuous state information from each of that entity's mentions. Separate LSTMs integrate entity-location information into this process. These continuous components then produce potentials for a sequential CRF tagging layer, which predicts discrete entity states. The CRF's problem-specific tag scheme, along with transition constraints, ensures that the model's predictions of these observed entity properties are structurally coherent. For example, in procedural texts, this involves ensuring existence before destruction and unique creation and destruction points. Because we use global inference, identifying implicit event creation or destruction is made easier, since the model resolves conflicts among competing time steps and chooses the best time step for these events during sequence prediction.

Past approaches in the literature have typically been  end-to-end continuous task specific frameworks \cite{henaff2016tracking,bosselut2018simulating}, sometimes for tasks that are simpler and more synthetic \cite{weston2015towards}, or continuous entity-centric neural language models \cite{N18-1204, D17-1195}. For process understanding specifically, past work has effectively captured global information \cite{dalvi2018tracking} and temporal characteristics \cite{das2018building}. However, these models do not leverage the structure constraints of the problem, or only handle them heuristically \cite{tandon2018reasoning}. We find that our model outperforms these past approaches on the \textsc{ProPara} dataset of \citet{dalvi2018tracking} with a significant boost in questions concerning entity state, regardless of the location.


\section{Model}
We propose a structured neural model for the process paragraph comprehension task of \citet{dalvi2018tracking}. 
An example from their dataset is shown in Figure~\ref{fig:grid}. It consists of annotation over a process paragraph $\mathbf{w} = \{w_i\}_{i=1}^P$ of $P$ tokens described by a sequence of $T$ sentences $\mathbf{s}=\{\mathbf{s}_t\}_{t=1}^T$. A pre-specified set of entities $E = \{e_k\}_{k=1}^{m}$ is given as well. For each entity, gold annotation is provided consisting of the state (\textsc{Exists}, \textsc{Moves}, etc.) and location (\emph{soil}, \emph{leaf}) after each sentence. From this information, a set of questions about the process can be answered deterministically as outlined in \citet{tandon2018reasoning}.


Our model, as depicted in Fig. \ref{fig:grid},  consists of two core modules: (i) state tracking, and (ii) location tracking. We follow past work on neural CRFs \cite{collobert2011natural,durrett-klein:2015:ACL-IJCNLP,N16-1030}, leveraging continuous LSTMs to distill information and a discrete CRF layer for prediction.

\subsection{State Tracking}

This part of the model is charged with modeling each entity's state over time. Our model places a distribution over state sequences $\mathbf{y}$ given a passage $\mathbf{w}$ and an entity $e$: $P(\mathbf{y}|\mathbf{w},e)$.

\paragraph{Contextual Embeddings} Our model first computes contextual embeddings for each word in the paragraph using a single layered bidirectional LSTM. Each token word $w_i$ is encoded as a vector $\vec{x}_i = [emb(w_i);v_i]$ which serves as input to the LSTM. Here, $emb(w_i) \in \mathbb{R}^{d_1}$  is an embedding for the word produced by either pre-trained GloVe \cite{pennington2014glove} or ELMo \cite{N18-1202} embeddings and $v_i$ is a scalar binary indicator if the current word is a verb. We denote by $\vec{h}_i = \mathrm{LSTM}([\vec{x}_i])$ the LSTM's output for the $i$th token in $\vec{w}$.

\paragraph{Entity Tracking LSTM} To track entities across sentences for state changes, we use another task specific bidirectional LSTM on top of the base LSTM which operates at the sentence level. The aim of this BiLSTM is to get a continuous representation of the entity's state at each time step, since not all time steps mention that entity. This representation can capture long-range information about the entity's state which may not be summarized in the discrete representation.

For a fixed entity $e$ and each sentence $\vec{s}_t$ in the paragraph, the input to the entity tracking LSTM is the contextual embedding of the \textit{mention location}\footnote{We use \emph{mention} location to differentiate these from the physical entity locations present in this QA domain.} of the entity $e$ in $\vec{s}_t$, or a mask vector when the entity isn't present in $\vec{s}_t$. Let $\vec{x}_{t}^e$ denote the representation of entity $e$ in sentence $t$. Then

\begin{equation}
 \vec{x}_{t}^{e}=\left\{
  \begin{array}{@{}ll@{}}
     [\vec{h}_{t}^e; \vec{h}_{t}^v], & \text{if}\ e \in \vec{s}_t \\
   \textrm{zero vector}, & \text{otherwise}
  \end{array}\right.
\end{equation} 
where $\vec{h}_{t}^e$ and $\vec{h}_{t}^v$ denote the contextual embeddings of the entity and the associated verb, respectively, from the base BiLSTM. In case of multiple tokens, a mean pooling over the token representations is used. Here, the information about verb is extracted using POS tags from an off-the-shelf POS tagger.
The entity tracking LSTM then produces representations $\widetilde{\vec{h}}_{t}^e = \mathrm{LSTM}([\vec{x}_{t}^e])$.

\paragraph{Neural CRF} We use the output of the entity tracking BiLSTM to generate emission potentials for each tag in our possible tag set at each time step $t$:
\begin{equation}
        \phi(y_t,t,\mathbf{w},e) = \vec{W}_{y_t}^T\widetilde{\vec{h}}_{t}^e
\end{equation}
where $\mathbf{W}$ is a learnable parameter matrix. For the specific case of entity tracking, we propose a 6 tag scheme where the tags are as follows:
\begin{table}[h]
\small
\begin{center}
\begin{tabular}{c|l}
\toprule
\bf Tags & \bf Description \\
\midrule
 $O_B, O_A$ & None state before and after existence, resp. \\
 $C, D$  & Creation and destruction event for entity, resp. \\
 $E$  & Exists in the process without any state change \\
$M$  & Entity moves from $loc_a$ to $loc_b$ \\
\bottomrule
\end{tabular}
\end{center}
\caption{\label{tag-scheme} Proposed tag scheme for the neural CRF based model for entity tracking.}
\end{table} \\
Additionally, we train a transition matrix to get transition potentials between tags which we denote by $\psi(y_{i-1}, y_i)$ and two extra tags:  $\bf{\langle START\rangle}$ and  $\bf{\langle STOP\rangle}$. Finally, for a tag sequence $\mathbf{y}$, we get the probability as:

\begin{equation}
        P(\mathbf{y}|\mathbf{w},e) \propto \exp\Big(\sum_{i=0}^T \phi(y_i,i,\mathbf{w},e) +  \psi(y_{i-1}, y_i)\Big)
\end{equation}

\subsection{Location Tracking}
To complement entity's state changes with the change in physical location of the entity, we use a separate recurrent module to predict the locations. Given a set of potential locations $L = (l_1,l_2,\ldots,l_n)$, where each $l_j \in L$ is a continuous span in $\vec{w}$, the location predictor outputs a distribution for a passage $\vec{w}$ and entity $e$, at a given time step $t$ as $P(l|\vec{w},e,t)$.


\paragraph{Identifying potential locations} Instead of considering all the spans of text as candidates for potential locations, we systematically reduce the set of locations by utilizing the part of speech (POS) tags of the tokens, whereby extracting all the maximal \textit{noun} and \textit{noun + adjective} spans as potential \textit{physical} location spans. Thus, using an off-the-shelf POS tagger, we get a set $L=(l_1,l_2,\ldots,l_n)$ of potential locations for each $\vec{w}$. These heuristics lead to a $85\%$ recall classifier for locations which are not null or unk.\footnote{Major non-matching cases include long phrases like ``deep in the earth'', ``side of the fault line'', and ``area of high elevation'' where the heuristics picks ``earth'', ``fault line'', and ``area'', respectively.}

\paragraph{Location Tracking LSTM} For a given location $l$ and an entity $e$, we take the mean of the hidden representations of tokens in the span of $l$ in $\vec{s}_t$ (or else a mask vector) analogous to the input for entity state tracking LSTM, concatenating it with the mention location of the entity $e$ in  $\vec{s}_t$, as input for time-step $t$ for the tracking this entity-location pair with $ \widetilde{\vec{h}}_{t}^{e,l} = \mathrm{LSTM}\left([\vec{x}^{e,l}_t]\right)$. Fig. \ref{fig:grid} shows an example where we instantiate location tracking LSTMs for each pair of entity $e$ and potential location $l$. In the example, $e \in \{water, CO2, sugar\}$ and $l \in \{soil, leaf\}$.
\begin{table*}[t!]
\small
\centering
\begin{tabular}{r|ccc|cc|ccc}
\toprule
  \multirow{2}{*}{\bf Model} & \multicolumn{5}{c|}{Task-1} &  \multicolumn{3}{c}{Task-2}\\
\cmidrule{2-9}
  & \bf Cat-1 & \bf Cat-2 & \bf Cat-3 & \bf Macro-Avg & \bf Micro-Avg & \bf Precision & \bf Recall & \ \bf  $F_1$ \\
   \midrule
    EntNet \cite{henaff2016tracking} & 51.62 & 18.83 & 7.77 & 26.07 & 25.96 & 50.2 & 33.5 & 40.2\\  QRN \cite{seo2016query} & 52.37 & 15.51 & 10.92 & 26.26 & 26.49 & 55.5 & 31.3 & 40.0\\ 
   ProGlobal \cite{dalvi2018tracking}& 62.95 & 36.39 & 35.90 & 45.08 & 45.37 & 46.7 & 52.4 & 49.4 \\  
    ProStruct \cite{tandon2018reasoning}& - &  - & - & - & - & 74.2 & 42.1 & 53.75 \\ 
    KG-MRC \cite{das2018building} & 62.86 &  40.00 & 38.23 & 47.03 & 46.62 & 64.52 & 50.68 & 56.77 \\ 
    \midrule
   This work: NCET & 70.55 & 44.57 & 41.34 & 52.15 & 52.31 & 64.2 & 53.9 & 58.6 \\
    This work: NCET + ELMo & 73.68 & 47.09 & 41.03 & 53.93 & 53.97 & 67.1 & 58.5 & 62.5  \\
    \bottomrule
\end{tabular}
\caption{\label{tab:task12}Results on the sentence-level (Task-1)  and document-level (Task-2)  evaluation task of the \textsc{ProPara} dataset on the test set. Our proposed CRF-based model achieves  state of the art results on both the tasks compared to the previous work in \cite{das2018building}. Incorporating ELMo further improves the performance for the state tracking module, as we see from the gains in Cat-1 and Cat-2.
  }
\end{table*}

\paragraph{Softmax over Location Potentials} The output of the location tracking LSTM is then used to generate potentials by for each entity $e$ and location $l$ pair for a time step $t$. Taking softmax over the potentials gives us a probability distribution over the locations $l$ at that time step $t$ for that entity $e$: $  p_{t}^{e,l} = \mathrm{softmax}(\mathbf{w}^T_{loc}{\widetilde{\vec{h}}_{t}^{e,l}})$

\subsection{Learning and Inference}
The full model is trained end-to-end by minimizing the negative log likelihood of the gold state tag sequence for each entity and process paragraph pair. The location predictor is only trained to make predictions when the gold location is defined for that entity in the dataset (i.e., the entity exists).

At inference time, we perform a global state change inference coupled with location prediction in a pipelined fashion. First, we use the state tracking module of the proposed model to predict the state change sequence with the maximum score using Viterbi decoding. Subsequently, we predict locations where the predicted tag is either \textit{create} or \textit{move}, which is sufficient to identify the object's location at all times since these are the only points where it can change.

\section{Experiments}
We evaluate the performance of the proposed model on the two comprehension tasks of the \textsc{ProPara} dataset \cite{dalvi2018tracking}. This dataset consists of 488 crowdsourced real world process paragraphs about 183 distinct topics in the science genre. The names of the participating entities and their existence spans are identified by expert annotators. Finally, crowd workers label locations of participant entities at each time step (sentence). The final data consists of 3.3k sentence with an average of 6.7 sentences and 4.17 entities per process paragraph. We compare our model, the Neural CRF Entity Tracking (NCET) model, with benchmark systems from past work.

\subsection{Task 1: Sentence Level}

This comprehension task concerns answering 10 fine grained sentence level templated questions grouped into three categories: (\textbf{Cat-1}) Is $e$ Created (Moved, Destroyed) in the process (yes/no for each)? (\textbf{Cat-2}) When was $e$ Created (Moved, Destroyed)? (\textbf{Cat-3}) Where was $e$ Created, (Moved from/to, Destroyed)? The ground truth for these questions were extracted by the application of simple rules to the annotated location state data. Note that Cat-1 and Cat-2 can be answered from our state-tracking model alone, and only Cat-3 involves location.

As shown in Table \ref{tab:task12}, our model using GloVe achieves state of the art performance on the test set. The performance gain is attributed to the gains in Cat-1 and Cat-2 ($7.69\%$ and $4.57\%$ absolute), owing to the structural constraints imposed by the CRF layer.  The gain in Cat-3 is relatively lower as it is the only sub-task involving location tracking. Additionally, using the frozen ELMo embedding the performance further improves with major improvements in Cat-1 and Cat-2.

\subsection{Task 2: Document Level}

 The document level evaluation tries to capture a more global context where the templated\footnote{Inputs refer to the entities which existed prior to the process and are destroyed during it. Outputs refer to the entities which get created in the process without subsequent destruction. Conversion refers to the simultaneous event which involves creation of some entities coupled with destruction of others.} questions set forth concern about the whole paragraph structure: (i) What are the inputs to the process? (ii) What are the outputs of the process? (iii) What conversions occur, when and where? (iv) What movements occur, when and where? Table \ref{tab:task12} shows the performance of the model on this task. We achieve state of the art results with a $F_1$ of $58.6$.

\subsection{Model Ablations}
We now examine the performance of the model by comparing its variants along two different dimensions: (i) modifying the structural constraints for the CRF layer, and (ii) making changes to the continuous entity tracking.
\paragraph{Discrete Structural Constraints} We experiment with two new tag schemes: (i) $tag_1: O_A=O_B$, and (ii) $tag_2:O_A=E=O_B$. As shown in Table \ref{ablations}, the proposed 6 tag scheme outperforms the simpler tag schemes indicating that the model is able to gain more from a better structural annotation. Additionally, we experiment with removing the transition features from our CRF layer, though we still use structural constraints. Taken together, these results show that carefully capturing the domain constraints in how entities change over time is an important factor in our model.
\paragraph{Continuous Entity Tracking} To evaluate the importance of different modules in our continuous entity tracking model, we experiment with (i) removing the verb information, and (ii) taking attention-based input for the entity tracking LSTM instead of the entity-mention information. This way instead of giving a hard attention by focusing exactly on the entity, we let the model learn soft attention across the tokens for each time-step. The model can now learn to look anywhere in a sentence for entity information, but is not given prior knowledge of how to do so. As shown, using attention-based input for entity tracking performs substantially worse, indicating the structural importance of passing the mask vector.

\begin{table}[t]
\small
\begin{center}
\begin{tabular}{l|c|c|c|c|c}
\toprule
\bf Model & C-1 & C-2 & C-3 &\bf Mac. & \bf Mic. \\ \midrule
NCET & 72.27 & 46.08 & 40.82 & 53.06 & 53.13 \\
\midrule
 Tag Set 1 & 71.53 & 41.89 & 41.42 & 51.61 & 51.94 \\
 Tag Set 2 & 71.97 & 41.85 & 39.71 & 51.18 & 51.43\\
 No trans. & 71.68 & 44.22 & 40.38 & 52.09 & 52.24 \\
\midrule
 No verb & 73.16 & 42.58 & 41.85 & 52.53 & 52.85 \\
 Attn. & 61.69 & 22.80 & 36.44 & 40.31 & 41.38 \\
\bottomrule
\end{tabular}
\end{center}
\caption{\label{ablations} Ablation studies for the proposed architecture.}
\end{table}

\section{Conclusion} In this paper, we present a structured architecture for entity tracking which leverages both the discrete and continuous characterization of the entity evolution. We use a neural CRF approach to model our discrete constraints while tracking entities and locations recurrently. Our model achieves state of the art results on the \textsc{ProPara} dataset.

\section*{Acknowledgments}
This work was partially supported by NSF Grant IIS-1814522, NSF Grant SHF-1762299, a Bloomberg Data Science Grant, and an equipment grant from NVIDIA. The authors acknowledge the Texas Advanced Computing Center (TACC) at The University of Texas at Austin for providing HPC resources used to conduct this research. Thanks to the anonymous reviewers for their helpful comments.
\bibliography{naaclhlt2019}
\bibliographystyle{acl_natbib}

\end{document}